\documentclass[11pt]{article}
\usepackage{syntaxfest2021}
\usepackage{times}
\usepackage{url}
\usepackage{latexsym}

\usepackage{graphicx}
\usepackage{caption}
\usepackage{graphics}
\usepackage[T1]{fontenc}
\usepackage[utf8]{inputenc}
\usepackage{microtype}

\usepackage{subcaption}

\usepackage{todonotes}
\usepackage{tikz-dependency}
\usepackage{booktabs}
\usepackage{gb4e}
\usepackage{tipa}

\syntaxfestfinalcopy 

\title{Dravidian language family through Universal Dependencies lens}

\author{Taraka Rama \\
   Independent Researcher \\ 
  {\tt } \\\And
  Sowmya Vajjala \\
  National Research Council, Canada\\
  {\tt sowmya.vajjala@nrc-cnrc.gc.ca} \\}

\date{}

\begin{document}
\maketitle
\begin{abstract}
The Universal Dependencies (UD) project aims to create a cross-linguistically consistent dependency annotation for multiple languages, to facilitate multilingual NLP. It currently supports 114 languages. Dravidian languages are  spoken by over 200 million people across the word, and yet there are only two languages from this family in UD. This paper examines some of the morphological and syntactic features of Dravidian languages and explores how they can be annotated in the UD framework.
\end{abstract}

\section{Introduction}
Treebanks are one of the basic resources needed for developing tools for natural language processing. However, developing a treebank requires significant annotation effort and early NLP research featured large treebank projects such as Penn Treebank \cite{Marcus.Santorini.ea-93} and Prague Dependency Treebank \cite{Bohomova.Haji.ea-03} each following their own annotation schema. Recently, the Universal Dependencies (UD)\footnote{The latest released version \url{http://hdl.handle.net/11234/1-3687} has 202 treebanks in 114 languages.} became a popular framework to create cross-linguistically consistent dependency treebanks. Such consistent dependency annotations are also useful for language typology and multilingual NLP studies. 

\begin{figure}[!ht]
\begin{subfigure}[b]{0.45\textwidth}
    \centering

    \includegraphics[width=0.7\textwidth,height=0.65\textwidth]{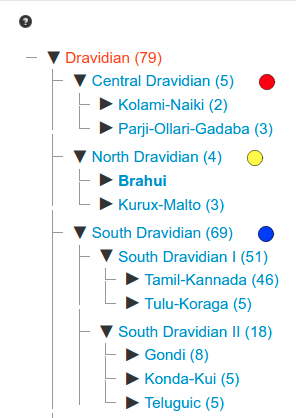}
   
    \caption{Dravidian family according to Glottolog.}
     \label{fig:DravGlot}

\end{subfigure}
\hfill
\begin{subfigure}[b]{0.45\textwidth}
    \centering

    \includegraphics[width=0.7\textwidth,height=0.8\textwidth]{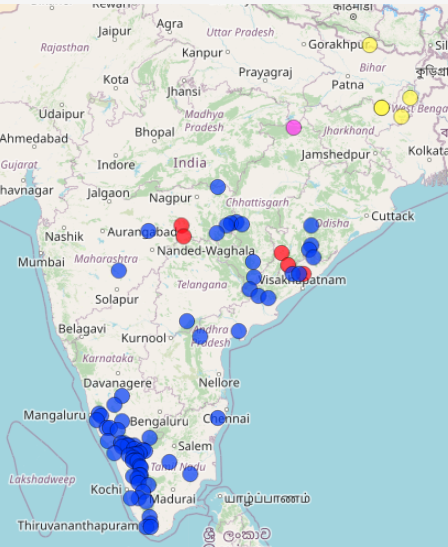}
    \caption{Map of the Dravidian languages with colors showing the subgroups.}
    \label{fig:DravMap}
\end{subfigure}
    \caption{Dravidian Languages' subgrouping and geographical distribution.}
\end{figure}

The Dravidian family consists of about 80\footnote{\url{https://glottolog.org/resource/languoid/id/drav1251}} language varieties (languages and dialects) that are spoken by more than 220 million people across the world. Until now, only 2 of the 114 languages represented in UD 2.8 belong to the Dravidian family---Tamil and Telugu. While there are treebanks for some of the major Dravidian languages in the P\={a}\d{n}inian \textit{k\={a}raka} framework \cite{Rao.MuraliKrishna.ea-14}, there are no computational tools to convert or compare the P\={a}\d{n}inian treebanks with the treebanks annotated in the UD framework. In this paper, we describe the characteristics of Dravidian languages with representative examples from a language from each of the four major branches (Figure \ref{fig:DravGlot} shows the phylogenetic tree) as representative examples, and discuss the UD framework annotation effort for this language family. 

This paper is organized as follows: We introduce the Dravidian language family and its salient features, and survey relevant literature in the section that follows (\S  \ref{sec:dlintro}). We will note our observations about some specific issues related to word tokenization including lemmatization (\S ~\ref{sec:token}), annotating morphology (\S ~\ref{sec:morph}), part-of-speech tagging (\S ~\ref{sec:pos}),  and syntax (\S ~\ref{sec:syntax}) for Dravidian languages in the sections that follow, comparing how similar issues were addressed in UD, where possible. Finally, Section \ref{sec:disc} summarizes the paper's main points of discussion.

\section{Dravidian Languages and Treebanks}
\label{sec:dlintro}
The Dravidian language family consists of around 80 language varieties, and is the \textit{sixth} largest language family in the world in terms of the number of speakers. Dravidian languages are primarily spoken in South Asia, concentrated around South and Central India (Figure \ref{fig:DravMap}). In addition, the languages are also spoken in many other parts of the world, through diaspora. Of these, four languages -- Telugu, Tamil, Kannada, and Malayalam -- are among the official languages of India. All the four languages have their own writing systems and a long standing (written) literary tradition. The Dravidian language family has four subgroups: South I, South II, Central, and North Dravidian (Figure \ref{fig:DravGlot}). The South Dravidian I subgroup consists of three of the four major literary Dravidian languages Tamil, Kannada, and Malayalam. We take Tamil as a representative language for this group. Telugu belongs to the South Dravidian II branch. We will analyze examples from Gadaba and Brahui languages that represent Central and North Dravidian subgroups respectively.

\paragraph{Major family features} Dravidian languages are agglutinative languages where suffixation and compounding are used instead of prefixes and infixes to express grammatical relations. Reduplication and echo-words are frequently seen in Dravidian languages, as is commonly seen in other language families of this linguistic area \cite{emeneau1956india}. The major word classes in Dravidian languages are: nominals (nouns, pronouns, numerals, adverbs of time and place), adjectives, verbs, adverbs, and clitics \cite[ch 2]{Krishnamurthi-03}. Negation is typically conveyed through verbal morphosyntax. Articles are not present as a separate word class, and their function is conveyed through other grammatical means \cite[ch 1]{Steever-19}. The languages are relatively free word order, with Subject-Object-Verb being the dominant word order. Inclination towards pro-drop is an important feature of all languages of this family. Most Dravidian languages have a clusivity distinction. Dative subject constructions are a common phenomenon. Relative clauses typically precede the head noun. The passive voice is rarely used. 

\subsection{Dravidian Treebanks}
Dravidian languages are under-represented in treebanking projects. All the research in this direction has focused on the P\={a}\d{n}inian framework for dependency annotations \cite{Begum.Husain.ea-08}. \newcite{Husain-09} and \newcite{Husain.Mannem.ea-10} describe a Telugu dependency treebank in P\={a}\d{n}inian framework, developed for a shared task on dependency parsing, which is not available under an open license. In the recent past, \newcite{Tandon.Sharma-17} described freely available treebanks for Indian languages in the same framework, that consisted of three Dravidian languages -- Telugu, Tamil, and Kannada. \newcite{Nallani.Shrivastava.ea-20} describe an enhanced Telugu dependency treebank building on the previous work in the P\={a}\d{n}inian framework. \newcite{Rao.MuraliKrishna.ea-14} described annotation guidelines for Kannada using P\={a}\d{n}inian framework. \newcite{Ramasamy-12} developed a Prague dependency style treebank for Tamil which is different from the P\={a}\d{n}inian framework based treebanks for the Dravidian languages.

\paragraph{UD and Dravidian Languages} Within the universal dependencies framework, Dravidian languages are among the least represented groups. Small treebanks exist for two major Dravidian languages: (1) Tamil \cite{Ramasamy-12}, converted from Prague style dependencies, approximately 500 sentences, (2) Telugu \cite{Rama.Vajjala-18}, developed from scratch based on a Telugu grammar book examples, about 1300 sentences. While there is some research on converting Hindi P\={a}\d{n}inian treebank to Universal Dependencies in the past \cite{Tandon.Chaudhry.ea-16}, we are not aware of any such conversion effort for the Dravidian languages. 

Given this background, there is a need for Dravidian language treebanks, to increase the diversity of UD and develop NLP resources, and support linguistic typlogy research. Considering that there is a large amount of recent research into developing treebanks from scratch for very under-resourced (and sometimes, endangered) languages such as Albanian \cite{Toska.Nivre.ea-20} and St. Lawrence Island Yupik \cite{Park.Lane.ea-21}, we believe that there should be efforts to build treebanks for various Dravidian languages as well.
Past research \cite{Bharathi.Sangal.ea-02,Begum.Husain.ea-08} described few challenges in annotation efforts for Indian languages, although the discussion entirely focused on Hindi. 
To our knowledge, this is the first paper to specifically address Dravidian language issues in the context of universal dependencies treebanking.

\paragraph{Notes on transcription} Following the South Asian transcription system employed in Dravidian scholarship, a bar under a consonant \b{t} represents alvelolar sound, a dot under a consonant \d{t} represents a retroflex sound. Tamil retroflex approximant \textipa{\:R} is written as \d{z} and a alveolar tap \textipa{R} through \b{r}. Vowel length is shown with a macron \={a}.

\section{Word Tokenization/Lemmatization}
\label{sec:token}

Dravidian languages are agglutinative in nature and inflection is a widely seen phenomenon. It is common to see multi-word tokens in a typical sentence or have single token sentences with dropped subjects. UD framework supports an extension of the original CONLL-X annotation scheme, to annotate such multi-word tokens, where words are indexed with integers, and multi-word tokens are indexed with integer ranges. While this seems to be a common annotation procedure with $67/114$ languages in UD 2.8 using it, \newcite{Park.Lane.ea-21} recently argued for a morpheme level annotation for highly inflected languages which might be a possible direction for annotating Dravidian language treebanks in UD.

\begin{figure}[ht!]
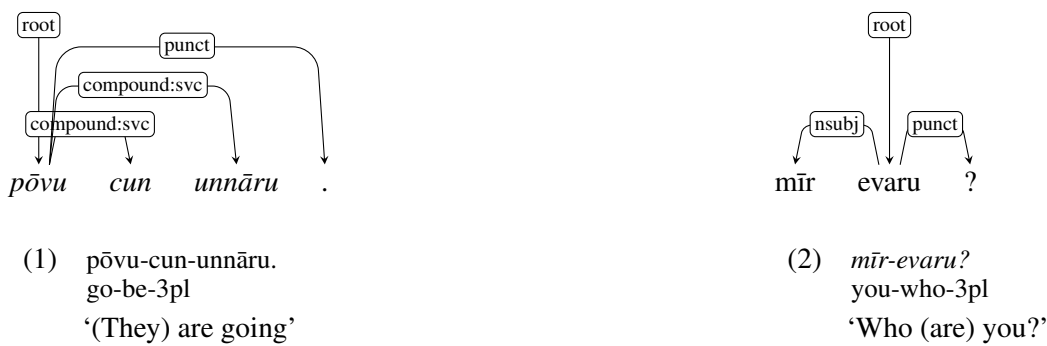


\begin{subfigure}[b]{0.6\textwidth}

\begin{dependency}[theme = default]
   \begin{deptext}[column sep=1em]
\textit{p\={o}vu} \& \textit{cun} \& \textit{unn\={a}ru} \& . \\
   \end{deptext}
   \deproot{1}{root}
   \depedge{1}{2}{compound:svc}
   \depedge{1}{3}{compound:svc}
   \depedge{1}{4}{punct}
   \end{dependency}

 \begin{exe}
\ex 
\gll \noindent
p\={o}vu-cun-unn\={a}ru. \\
go-be-3pl \\
\glt `(They) are going'
  \end{exe}
\end{subfigure}
\quad
\begin{subfigure}[b]{0.4\textwidth}
\begin{dependency}[theme = default]
   \begin{deptext}[column sep=1em]
m\={\i}r \& evaru \& ? \\
   \end{deptext}
   \deproot{2}{root}
   \depedge{2}{1}{nsubj}
   \depedge{2}{3}{punct}
   \end{dependency}

\begin{exe}
\ex 
\gll \noindent
\textit{m\={\i}r-evaru?} \\
you-who-3pl\\
\glt   `Who (are) you?'
 \end{exe}
\end{subfigure}
\caption{Token and Morpheme level annotation}
\label{fig:tokens}
\end{figure}

For example, consider the following two single token sentences from Telugu, shown in Figure \ref{fig:tokens}. The first one is a compound/serial verb with dropped subject from Old Telugu (ancestral language from the fifteenth century), along with its contemporary variant \cite[p. 200]{Steever-19}. The second sentence is a question with a non-overt copula, where the two words in the sentence are fused into one. A dependency annotation following token level convention in these cases will only show a \texttt{PUNCT} relation between the root (which is the single token) and the punctuation marker. A morpheme level annotation may be more suited for such case, as shown in Figure~\ref{fig:tokens}.



Although the existing Dravidian language treebanks in UD do not have such morpheme level annotation, it will be worthwhile to weigh this aspect while developing new Dravidian treebanks in future. However, it is worth noting that there are no ready to use, easily accessible tokenizers/morph analyzers/lemmatizers for most of these languages as of now. The Unimorph \cite{Kirov.Cotterell.ea-18} project, which currently supports morphological analyzers for three Dravidian languages (Telugu, Tamil, Kannada), can be used as a starting point for this purpose.

\section{Morphological Features}
\label{sec:morph}
Morphological specification of a word in UD consists of three levels: lemma, morphological features, and POS tags. We discussed the lemma annotation in the previous section. In this section, we discuss some common morphological features in Dravidian languages and how they could be addressed within the UD framework.  Dravidian languages have two major word classes: nouns and verbs \cite{Krishnamurthi-03}. Nouns possess number/gender and are inflected for case, along with pronouns and numerals, which are considered nominals in the Dravidian comparative scholarship. Adverbs of time and place can also inflect for case. Verbs show agreement with number, gender and person of its agent and are inflected for tense, aspect, mood and modality. Case relations are expressed either by bound morphemes or by post positions. Negation is typically conveyed through verbal morphosyntax i.e., verbs have both positive and negative forms. The following subsections discuss some specific morphological features in the context of Dravidian languages and UD. We mapped the morphological features in the comparative handbooks to the UD schema.



\subsection{Clitics} Clitic is a bound morpheme that is phonologically dependent on a host, but syntactically independent \cite[ch 5]{Velupillai-12}. In UD, clitics are language specific features, where some languages (e.g., Polish) mark it as a Yes/No feature, while others (e.g., Finnish and Skolt-Sami) use feature values that express the function of the clitic (e.g., question, focus). In Dravidian languages, clitics perform various functions, such as: interrogative (e.g., Tamil: \textit{ava\b{n}\={a}} - ``is it him?''), emphasis (e.g., Telugu: \textit{\textbf{ad\={e}} n\={a} pustakam} - ``\textbf{That} is my book''), and conjunction (Tamil: \textit{kamala\textbf{v\=o} car\=oja\textbf{v\=o}} - ``either Kamala \textbf{or} Saroja''). Any constituent can be turned into a question using a clitic, as illustrated by the following Tamil examples in Figure~\ref{fig:cli1}.  

\begin{figure}[htb!]
\small
\centering
\begin{subfigure}[b]{0.5\textwidth}
\begin{exe}
    \ex \gll \noindent
    avan n\={e}\b{r}\b{r}u va-nt-\={a}\b{n}-\={a}?\\
    that.man-Nom    yesterday   come-Past-3|Sing|Masc|Int\\
\glt `Did he come yesterday?\\
\end{exe}
\end{subfigure}
~
\begin{subfigure}[b]{0.45\textwidth}
\begin{exe}
    \ex \gll \noindent
    ka\d{n}\d{n}a\b{n} eppo\d{z}utu varu-ki\b{r}-\={a}\b{n}?\\
    Kannan-Nom when come-Pres-3|Sing|Masc\\
\glt `When is Kannan coming?\\
\end{exe}
\end{subfigure}
\caption{Interrogative Clitics in Tamil}
\label{fig:cli1}
\end{figure}

Additionally, Brahui has first and second person clitic pronouns, which have a genitive meaning when suffixed to a nominal and dative-accusative when suffixed to a verb \cite[p. 397]{Steever-19}, as shown in Figure~\ref{fig:cli2}.
\begin{figure}[htb!]
\centering
\small
\begin{subfigure}[t]{0.4\textwidth}
\begin{exe}
\ex \gll 
b\={a}va-ta	p\={a}r\={e} \\
father-Gen say-Past-3|Sing\\
\glt`His/Her father said.'
\end{exe}
\end{subfigure}
\begin{subfigure}[t]{0.4\textwidth}
\begin{exe}
\ex \gll 
b\={a}va-t\={a} 	p\={a}r\={e}-t\={a}\\
father-Gen|Plur	say-Past-3|Plur|Acc \\
\glt`Their father said to them.'
\end{exe}
\end{subfigure}
\caption{Clitics in Brahui}
\label{fig:cli2}
\end{figure}

Considering these diverse functions, it is perhaps appropriate to mark the value of the clitic feature in Dravidian languages in terms of its function in the sentence, rather than a binary yes/no value. Sometimes, clitics may also appear as separate tokens, which we will discuss in the next section.  


\subsection{Compound Words}
Compunds might need a special treatment, as suggested by \newcite{Marneffe.Manning.ea-21}, since they can appear both with and without word boundary in Dravidian languages. When appearing with a word boundary, they can be related by a dependency relation (\texttt{compound}) along with a subtype (e.g., \texttt{redup, svc}) where needed. However, when they appear as a single token in a sentence, a language specific feature \textit{compound} may be used, as is currently being done with languages such as Sanskrit, Estonian, and Romanian.

\paragraph{Reduplication} is a frequently observed morphological process in South Asian languages, especially among the Indo-aryan and Dravidian language families. Echo words and onomatopoeia are also observed with reduplication \cite{Abbi-94}. These can be considered a special case of compound words. However, within or outside UD, there is no consistent scheme for annotating these phenomenon, and it is generally considered a language specific feature. While some treebanks annotated reduplication as a syntactic relation \textit{compound:redup} (e.g., Naija, Turkish, and Uyghur), others annotated it as a morphological feature Echo, taking a value ``Rdp'' or ``Ech'' with or without annotating the syntactic relation explicitly as a \texttt{compound} or \texttt{compound:redup} (e.g., Hindi). 

In the Hindi treebank \cite{Bhat.Bhatt.ea-17}, it was proposed that reduplication and echo words have their own POS tags. However, since these words play different syntactic roles in a sentence, it is more appropriate to consider reduplication as a morphological feature, and use the POS tags as per their role in the sentence. In the case of reduplication within a single word, we could annotate the morphological feature of the word indicating the same, and POS tag can follow the role of the word in the sentence. If the reduplicated compound appears as two words instead of one, they can be connected by the dependency relation compound, with its subtype \texttt{redup}.  This is illustrated in the Tamil example below. 

\begin{figure}
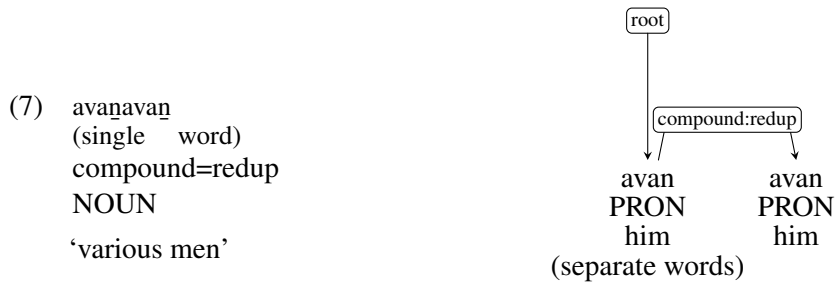

\begin{subfigure}[b]{0.45\textwidth}
\begin{exe}
    \ex \gll 
    ava\b{n}ava\b{n} \\
    (single word)\\
    compound=redup\\
    NOUN
    \glt`various men'\\
\end{exe}
\end{subfigure}
\begin{subfigure}[b]{0.45\textwidth}
\begin{dependency}[theme = default]
   \begin{deptext}[column sep=0em]
avan   \& avan \\ 
PRON \& PRON \\
him \& him \\
(separate words) \\
   \end{deptext}
   \deproot{1}{root}
   \depedge{1}{2}{compound:redup}
\end{dependency}
\end{subfigure}
\caption{Morphological analysis for Reduplication}
\label{fig:redup1}
\end{figure}

However, a challenging scenario for analysis comes with sentences where the reduplicated word also known as echo words such as \textit{puli-gili} `tigers and the like' \cite[p. 10]{emeneau1956india} where the first syllable in the previous word is replaced by \textit{gi-} in the Dravidian family. Consider the following Telugu sentence  from \newcite[p. 487]{Krishnamurthi-03}.

\begin{exe}
    \ex \gll \noindent
    v\={a}\d{d}i-ki \textbf{i}llu l\={e}du , \textbf{gi}llu l\={e}du .\\
    He-Acc    house not-VERB , thing-compound=redup not-VERB . \\    %

\glt`He has no house and nothing of that sort'
\end{exe}
Reduplicated words don't always have a meaning of their own (e.g., \textit{gillu} in this example). Considering that such sentence constructions are common in all several Dravidian languages, we need a standardized way of annotating the morphological features of such words.

\paragraph{Onomatopoeia} hasn't been discussed much in Universal Dependencies research, to our knowledge. While \newcite{Badmeva.Tyers-17} propose to treat them as interjections, \newcite{Vasquez.Aguirre.ea-18} propose to use the tag ``X'', and consider them as closed part of speech. However, there hasn't been any discussion on the morphological tagging of these words. We propose to use the POS tag and dependency relation for these words according to their syntactic function (e.g., adverb), and use a new morphological feature to indicate that it is an onomatopoeic word. Such words can also be reduplicative words in Dravidian languages. In that case, we can follow the same annotation procedure described earlier for compound words and reduplicative compounds. 




\section{Part-of-speech annotation}
\label{sec:pos}

Dravidian languages have two primary classes of words: Nominals and Verbals. Nominals in Dravidian include nouns, pronouns, numerals and time and place adverbs, and are the class of words that are inflected for \textit{case} \cite[ch. 6]{Krishnamurthi-03}. All nominals except adverbs are also inflected for person, gender and number. In this section, we discuss some of the POS tags, showing how the UD framework can be used for Dravidian language context. 

\paragraph{Numerals} In Dravidian languages, numerals can be inflected for case and can sometimes perform the function of an adjective or an adverb. Inflected numerals should be POS-tagged as per syntactic function, and not NUM, as long as they are not a part of a numeric expression. This is \textit{in agreement} with UD's recommendation to tag pronominal numerals and quantifiers as \texttt{DET}, instead of \texttt{NUM}. 
\paragraph{Adverbs of time and place} These typically behave like a noun morphologically in Dravidian languages, and are inflected for case. Hence, as  in the UD Telugu treebank \cite{Rama.Vajjala-18}, it would be appropriate to treat them as Nouns for Dravidian languages, marking the dependency relation according to its function in the sentence. 


\paragraph{Adpositions (ADP)} Postpositions are used in Dravidian languages to indicate case. They can also sometimes perform an adverbial function, to indicate temporal or location information, as nominal modifiers. Some languages, for example Brahui, also borrowed prepositions from its Indo-Aryan neighbor Baluchi \cite{Steever-19}. All of these should be tagged \texttt{ADP} in Dravidian languages. The following two examples illustrate the use of postpositions in Tamil.

\begin{exe}
    \ex \gll 
    avan k\={o}yil-ai p\={a}rttu p\={o}y iru-kkir-\={a}n\\
    that.man-Nom    temple-Acc  towards go-Sub  be-Pres-1|Sing\\
    PRON \hspace{2.5em} NOUN \hspace{1.2em}  ADP \hspace{0.7em}  VERB \hspace{0.7em}   VERB
    \glt `He has gone towards the temple'\\
\end{exe}


\paragraph{Quantifiers}  are treated as a sub-group of nominals by \newcite{Krishnamurthi-03}. However, UD framework includes (pronominal) quantifiers such as \textit{many, few} under the POS tag \texttt{DET}. Annotating as \texttt{DET} would maintain consistency across UD languages in the Dravidian treebanks. 


\subsection{Clitics} Clitics are treated as a separate part of speech class in the comparative Dravidan linguistic scholarship. When they appear as a separate token, clitics typically function as tag questions, as shown in the Telugu example below (Figure~\ref{fig:cli3}). There is no clear guideline on how to POS tag such items in UD. Since this is most likely a closed list of words in a language, we can perhaps annotate them as \texttt{PART}. The following three Telugu sentences  illustrate how clitics can appear as separate tokens in a sentence. 

\begin{figure}[htb!]
\begin{subfigure}[b]{0.4\textwidth}

\begin{exe}
\ex \gll \noindent
idi  m\={\i}  illu  \textbf{gad\={u}}?\\
This your house, isn't\\
DET PRON NOUN PART
\glt`This is your house, isn't it?' \\
\end{exe}
\end{subfigure}
~
\begin{subfigure}[b]{0.4\textwidth}
\begin{exe}
\ex \gll \noindent
idi \textbf{gad\={u}} m\={\i} illu?\\
This isn't your house \\
DET PART PRON NOUN
\glt`Isn't this your house?' \\
\end{exe}
\end{subfigure}
\caption{Clitics as separate words highlighted in bold.}
\label{fig:cli3}
\end{figure}

\subsection{Adverbs}
Onomatopoetic words are used as manner adverbs in many Dravidian languages without any inflection \cite[p 406]{Krishnamurthi-03}. Adverbs of manner can also come with a suffix. These phenomena are illustrated using Telugu and Tamil examples in Figure~\ref{fig:adv}. 
\begin{figure}[htb!]
\centering
\begin{subfigure}[b]{0.4\textwidth}
\begin{exe}
\ex \gll \noindent
nuvvu ga\d{d}a-ga\d{d}\={a} m\={a}\d{t}la\d{d}ut\={a}vu .\\
you fast-fast talk-Prog-2|Sing .\\
PRON ADV VERB PUNCT
\glt`You talk fast'. \\
\end{exe}
\end{subfigure}
\begin{subfigure}[b]{0.55\textwidth}
\begin{exe}
\ex \gll
onne vi\d{t}a ava v\={e}kam-a p\={e}cu-va .\\
you-Acc than she fast speak-3|Fem|Sing \\
PRON ADP PRON ADV VERB PUNCT
\glt`She speaks faster than you'\\
\end{exe}
\end{subfigure}
\caption{Adverbs in Telugu and Tamil}
\label{fig:adv}
\end{figure}




\subsection{Auxiliary Verbs (AUX)} Dravidian languages do not have a separate class of auxiliary verbs, and words that function as auxiliary verbs can also be the main verbs in other sentences. Hence they are not a closed class. This is illustrated in the following example from Telugu. 

 \begin{figure}[ht!]
 \small
\begin{subfigure}[b]{0.4\textwidth}
\begin{exe}
\ex \gll
meeru  in\d{t}il\={o}  un\d{d}an\d{d}i .\\
 you  at-home be-Hon-3|Plur .\\
 \glt `(You) stay at home'\\
\end{exe}
\end{subfigure}
~
\begin{subfigure}[b]{0.45\textwidth}
\begin{exe}
\ex \gll
aa  pustakam n\={a}ku  ivvan\d{d}i .\\
that  book  me-Case=Dat give-Hon-3|Plur .\\
\glt  `Give me that book'\\
\end{exe}
\end{subfigure}

\begin{subfigure}[b]{0.5\textwidth}
\centering
\begin{exe}
\ex \gll
nannu ikka\d{d}a  un\d{d}an-ivvan\d{d}i .\\
 I-Acc here stay-give .\\
\glt `Let me stay here'
\end{exe}
\end{subfigure}

\caption{Auxiliary Verbs in Telugu}
\label{fig:aux}
 \end{figure}










In these examples, the verb \textit{un\d{d}u} is a main verb in the first example, but is a supplementary verb in another. Hence, it may be appropriate to tag them as VERB and connect them through a compound relation. Further, it is not uncommon to see the two words appear as one compound word without a separator between them. Hence, it is appropriate to tag such words as VERB instead of AUX, and not use the AUX tag in Dravidian treebanks.

\subsection{Conjunction} 
\begin{figure}[!htb]
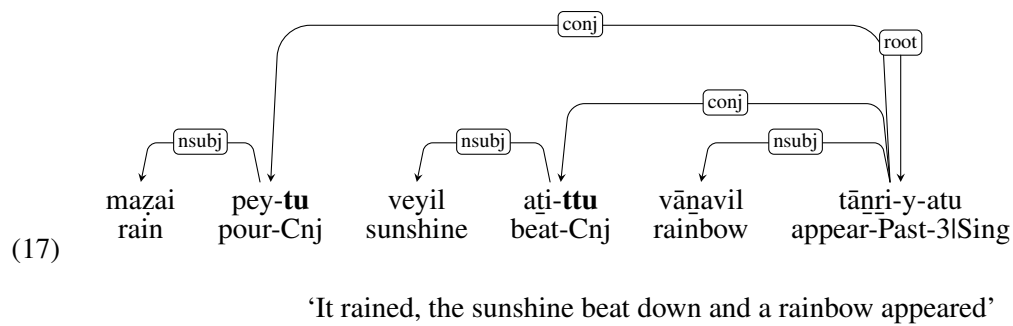

\centering
\begin{exe} \ex 
\begin{dependency}
\begin{deptext}[column sep=1em]
ma\d{z}ai \& pey-\textbf{tu} \& veyil \& a\b{t}i-\textbf{ttu} \& v\={a}\b{n}avil \& t\={a}\b{n}\b{r}i-y-atu\\
rain \& pour-Cnj \& sunshine \& beat-Cnj \& rainbow \& appear-Past-3|Sing\\   
\end{deptext}

\depedge{6}{4}{conj}
\depedge{6}{2}{conj}
\depedge{2}{1}{nsubj}
\depedge{4}{3}{nsubj}   
\depedge{6}{5}{nsubj} 
\deproot{6}{root}
\end{dependency}

\begin{center}
`It rained, the sunshine beat down and a rainbow appeared'
\end{center}
\end{exe}
\caption{Conjunctional suffices are in \textbf{bold}.}
\label{fig:Conjunctions}
\end{figure}

Dravidian languages typically indicate co-ordination through juxtaposition, instead of using an overt conjunction. Sometimes, it is also indicated by non-finite word forms. Thus, it is appropriate to use the POS tag reflecting the syntactic function of the word, as shown in the following Tamil sentence (Figure \ref{fig:Conjunctions}). 


While explicit conjunction words (e.g., \textit{mariyu} `and' in Telugu) can be tagged with a CCONJ/SCONJ tag,  But in most cases, since such usage is uncommon, it will be indicated through dependency relations. These are some of the issues we encountered while studying the POS tagging of Dravidian languages from the perspective of the UPOS tagset. The next section discusses some of the syntactic characteristics of these languages, and how they can be annotated in the UD framework.



\section{Syntactic Annotation}
\label{sec:syntax}

The dominant word order in the Dravidian languages is Sentence-Object-Verb (SOV) order, although the rich morphology allows a relatively free word order. Adpositions typically follow the noun, adverb precedes the verb, dependent clauses precede main clauses, genitives precede the nouns they modify and the main verbs precede auxiliaries \cite[Ch. 1]{Steever-19}. They are also pro-drop languages, and hence, a pronominal subject may be optionally deleted from the sentence, but its identity can be inferred from the word ending in most cases. Dravidian languages also feature non-verbal predicates. In this section, we will discuss some specific syntactic features of Dravidian languages within the UD framework.

\subsection{Free word order}
The rich nominal and verbal morphology in Dravidian languages allow a relatively free word order, and it is possible to change the unmarked word order of the constituents, keeping the verb at the end without effecting the meaning and perhaps the dependency structure of the sentence. This is illustrated in the Tamil sentences shown in Figure~\ref{fig:wordorder} below \cite[p. 425]{Krishnamurthi-03}.

\begin{figure}[htb!]

\begin{exe} \ex
\begin{dependency}[theme = default]
   \begin{deptext}[column sep=1em]
n\={e}\b{r}\b{r}u \& mantiri-avarka\d{l} \& ku\d{z}aint-ai-kku \& paric-ai \& ko\d{t}u-tt\={a}ru \&.\\
yesterday \& minister-Hon-Plur \& child-Dat \& prize-Acc \& give-past-3|Hon|Plur \& .\\
   \end{deptext} 
   \deproot{5}{root}
   \depedge{5}{1}{obl:tmod}
   \depedge{5}{2}{nsubj}
   \depedge{5}{3}{iobj}
   \depedge{5}{4}{obj}
   \depedge{5}{6}{punct}
\end{dependency}
\begin{center}
`Yesterday, the minister gave the child a prize.'    
\end{center}
\end{exe}

\begin{exe}
\ex
\begin{dependency}[theme = default]
   \begin{deptext}[column sep=1em]
   n\={e}\b{r}\b{r}u \& ku\d{z}aint-aikku \& paric-ai \& ko\b{t}u-tt\={a}ru \& \textbf{mantiri-avarkal} \& .\\
   \end{deptext}
   \deproot{5}{root}
   \depedge{5}{4}{acl:relcl}
   \depedge{4}{1}{obl:tmod}
   \depedge{4}{2}{iobj}
   \depedge{4}{3}{obj}   
   \depedge{5}{6}{punct}
   \end{dependency}
\begin{center}
`Yesterday, (it was) \textbf{the minister} (who) gave the child a prize.'
\end{center}
\end{exe}



   
 \begin{exe} \ex
 mantiri-avarkal ku\d{z}aint-ai-kku p.paric-ai k.ko\d{t}utt\d{a}ru \textbf{n\={e}\b{r}\b{r}u} .
\glt'\textbf{It was yesterday}, that the minister gave the child the prize.'
 \end{exe}
 \begin{exe} \ex
 n\={e}\b{r}\b{r}u mantiri-avarkal p.paric-ai k.ko\={t}utt\d{a}ru \textbf{ku\d{z}aint-ai.kku}  .
 \glt'Yesterday, (it was) \textbf{to the child} (that) the minister gave a prize..'
 \end{exe}
\caption{Free word order that allows shift in \textbf{focus} (shown in bold).}
\label{fig:wordorder}
\end{figure}
In this sentence, the adverb of time (which we treat as a nominal), the subject, and the direct and indirect objects can be shifted to any position keeping the verb in the final position. The dependency trees for examples (20) and (21) would be similar to (19) with the root being the word in focus. Cleft constructions can be formed by the use of clitics to shift focus. 



\subsection{Non-verbal predication}
In Dravidian languages, the predicate of a simple sentence can is not always a verb, and can be a nominal. This can also be observed in sentences that are questions. Figure \ref{fig:nvpred} illustrates two such sentences from Gadaba. 

\begin{figure}[htb!]
\centering
\begin{subfigure}[b]{0.45\textwidth}
\centering
\begin{exe} \ex 

   \begin{dependency}[theme = default]
   \begin{deptext}[column sep=0.5em]
\={o}\d{n} \&  kalgerte\d{n} \&  .\\
PRON \& PRON \&  PUNCT\\
He \& rich man \& .\\
   \end{deptext} 
   \deproot{2}{root}
   \depedge{2}{1}{nsubj}
   \depedge{2}{3}{punct}
\end{dependency}
\end{exe}
\begin{center}
`He is a rich man.'
\end{center}
\end{subfigure}
\centering
\begin{subfigure}[b]{0.5\textwidth}
\begin{exe} \ex 
   \begin{dependency}[theme = default]
   \begin{deptext}[column sep=0.5em]
\={\i} \&  kor \& \={e}yr-ne \& ?\\
PRON \& NOUN \& PRON \&  PUNCT\\
This \& fowl \& who-Case=Gen \& ?\\
   \end{deptext} 
   \deproot{3}{root}
   \depedge{2}{1}{det}
   \depedge{3}{2}{nsubj}
   \depedge{3}{4}{punct}
\end{dependency}
\begin{center}
`Whose fowl is this?    
\end{center}
 \end{exe}
\end{subfigure}
\caption{Non-verbal predication}
\label{fig:nvpred}
\end{figure}

\subsection{Dative subjects}
In dative subject constructions, the subject of some forms of verbs appears as a post-positional phrase with the nominal subject in the sentence-initial position in dative case \cite[p. 450]{Krishnamurthi-03}. All the Dravidian languages show dative constructions. Typically, the subject of a ``verb of emotion, sensation, cognition or possession appears in the dative case'' \cite[p. 29]{Steever-19}. This can be followed by a predicate represented either by a noun phrase or a verb phrase. UD's dependency annotation scheme does not have a relation to indicate such constructions. \newcite{Rama.Vajjala-18} used a relation \texttt{nsubj:nc} to indicate such relations (\texttt{nc} stands for non-canonical from the Persian treebank). It is perhaps appropriate to use either that or a new relation \texttt{nsubj:dat} for such constructions, when annotating dative subjects.

\begin{figure}[ht!]

\begin{subfigure}[b]{0.35\textwidth}
\begin{exe} \ex 
\begin{dependency}[theme = default]
   \begin{deptext}[column sep=.1cm]
avanukku  \& kopam \& va-nt-atu \& . \\
he-Dat \& anger \& got \& .\\
   \end{deptext}
   \deproot{3}{root}
   \depedge{3}{1}{nsubj:dat}
   \depedge{3}{2}{amod}
   \depedge{3}{4}{punct}
\end{dependency}
\begin{center}
`He is angry.'
\end{center}
\end{exe}
\caption{Tamil}
\end{subfigure}
~
 \begin{subfigure}[b]{0.4\textwidth}
 \begin{exe} \ex 
 \begin{dependency}[theme = default]
   \begin{deptext}[column sep=1em]
 bandas-ag-e \&[.1cm] ira	\&	mar	\&	assur \& .\\	
 man-one-Dat/Acc \& two \& son \& be-Past-3|Plur \& .\\
   \end{deptext}
   \deproot{4}{root}
   \depedge{4}{1}{nsubj:dat}
   \depedge{4}{3}{obj}
   \depedge{3}{2}{nummod}
   \depedge{4}{5}{punct}
 \end{dependency}
 \begin{center}
     `A man had two sons'
 \end{center}
\end{exe}
\caption{Brahui}
 \end{subfigure}
 \\
\begin{subfigure}[b]{0.4\textwidth}
\begin{exe} \ex 
\begin{dependency}[theme = default]
  \begin{deptext}[column sep=0.8em]
 w\={a}\d{d}iki \& oka \& ko\d{d}uku \& .\\
 He-Case=Acc \& one \& son \& .\\
  \end{deptext}
  \deproot{3}{root}
  \depedge{3}{1}{nsubj:dat}
  \depedge{3}{2}{nummod}
  \depedge{3}{4}{punct}
\end{dependency}
\begin{center}
  `He has a son.'  
\end{center}
\end{exe}
\caption{Telugu}
\end{subfigure}
~
\begin{subfigure}[b]{0.5\textwidth}
\begin{exe} \ex 
\begin{dependency}[theme = default]
  \begin{deptext}[column sep=0em]
anin	\&	santosam	\&	b\={e}te \& .\\	
I-Case=Dat \& happiness \& get.a.feeling-Past \& .\\
  \end{deptext}
  \deproot{3}{root}
  \depedge{3}{1}{nsubj:dat}
  \depedge{3}{2}{adjmod}
  \depedge{3}{4}{punct}
\end{dependency}
\begin{center}
    `I felt happy'
\end{center}
\end{exe}
\caption{Gadba}
\end{subfigure}
\caption{Dative Subjects}
\label{fig:datsub}
\end{figure}

The following examples (Figure \ref{fig:datsub}) from Tamil, Telugu, Brahui, and Gadaba respectively illustrate dative subject constructions in the Dravidian languages. While the first and last examples illustrate dative subject sentences with a verbal predicate, the other two show dative subject constructions with nominal predicates. In Gadaba, the subject (\textit{anin}) appears in an all purpose objective case, rather than a dedicated dative case, although it is considered a dative subject construction \cite[p. 370]{Steever-19}.



\subsection{Relative clauses}
\begin{figure}[htb!]
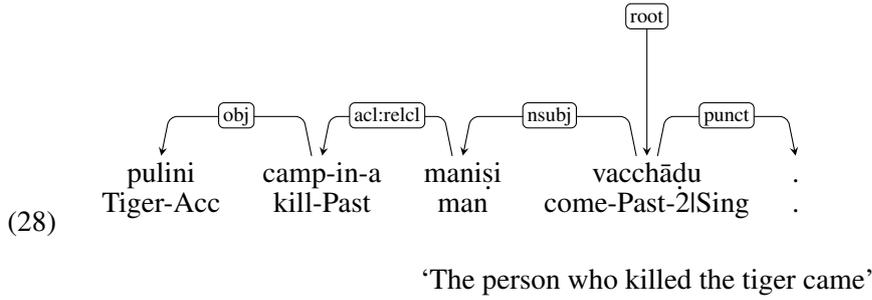

\centering
\begin{exe}
\ex
\begin{dependency}[theme = default]
  \begin{deptext}[column sep=1em]
pulini \& camp-in-a \& mani\d{s}i \& vacch\={a}\d{d}u \&.\\
Tiger-Acc \& kill-Past \& man \& come-Past-2|Sing \& .\\
  \end{deptext}
  \deproot{4}{root}
  \depedge{4}{5}{punct}
  \depedge{4}{3}{nsubj}
  \depedge{3}{2}{acl:relcl}
  \depedge{2}{1}{obj} 
\end{dependency}
\begin{center}
`The person who killed the tiger came'    
\end{center}
\end{exe}
\caption{Relative Clause in Telugu}
\label{fig:relcl}
\end{figure}

In Dravidian, relative clauses can be formed by replacing the verb of a sentence with a verbal adjective, instead of a separate relative pronoun,  The noun that is modified appears after the relative clause. The whole clause with this noun head becomes the clausal subject for the main verb. This phenomenon is illustrated in Figure~\ref{fig:relcl} for Telugu.

\subsection{Adverbial clauses}
There are three types of adverbial clauses in Dravidian \cite[p. 483]{Krishnamurthi-03}: a) a relative clause with an adverbial phrase of time or place and an adverbial head, b) a noun clause followed by post positions to indicate time and place and c) an embedded adverbial phrase with a manner adverbial as head. The following Telugu and Tamil examples 
from Figure \ref{fig:advcl} illustrate the use of adverbial clauses.
\begin{figure}[htb!]
 \begin{exe} 
 \ex
\begin{dependency}[theme = default]
   \begin{deptext}[column sep=.5em]
     atanu \& upany\={a}sam \& cebu-tun-na \& m\={u}\d{d}u \& gan\d{t}alu \& w\={a}na \& kuris-in-di \& .\\
     He \& lecture \& speak-Prog-Perf \& three \& hour-Plur \& rain \& rain-Past-3|Neut|Sing \& .\\
   \end{deptext}
   \deproot{7}{root}
   \depedge{7}{8}{punct}
   \depedge{7}{6}{nsubj}
   \depedge{7}{3}{advcl}
   \depedge{5}{4}{nummod}
   \depedge{3}{5}{obl:tmod}
   \depedge{3}{2}{obj}
   \depedge{3}{1}{nsubj}
\end{dependency}
\begin{center}
`It rained during the three hours he was lecturing.'    
 \end{center}
 \end{exe}
\begin{exe} \ex
\begin{dependency}[theme = default]
  \begin{deptext}[column sep=0em]
     ma\d{z}a\b{l}  \& pey-t-atu-kku \& appuram \& payir \& nanraka \& valar-nt-adu \& .\\
     rain-Nom  \& fall-Past-noml-dat \& after \& crops \& goodness-Adv \& grow-Past-3|Neut|Sing\\ 
  \end{deptext}
  \deproot{6}{root}
  \depedge{6}{7}{punct}
  \depedge{6}{5}{advmod}
  \depedge{6}{4}{nsubj}
  \depedge{6}{2}{advcl}
  \depedge{2}{3}{obl:tmod}
  \depedge{2}{1}{nsubj}
\end{dependency}
\begin{center}
`After it rained/rains, the crops grew/grow well'     
\end{center}
\end{exe}

\begin{exe}
\ex
\begin{dependency}[theme = default]
  \begin{deptext}[column sep=1em]
     v\={a}\d{d}u \& parige\d{t}\d{t}u-ko\d{n}\d{t}\={u} \& tondara-g\={a} \& wacc\={a}-\d{d}u \& .\\
     he \& run-Refl|Prog \& quick-Adv \& come-Past-3|M|Sing \& .\\
  \end{deptext}
  \deproot{4}{root}
  \depedge{4}{5}{punct}
  \depedge{4}{2}{advcl}
  \depedge{2}{3}{advmod}
  \depedge{4}{1}{nsubj}
\end{dependency}
\begin{center}
`He came running fast.'    
\end{center}

\end{exe}

\caption{Adverbial Clauses. Examples (29)  and (31) are Telugu and (30) is Tamil.}
\label{fig:advcl}
\end{figure}

\section{Discussion}
\label{sec:disc}
We discussed a few morphological and syntactic features of Dravidian languages in the context of Universal Dependencies framework, taking four languages, one from each of its subgroups -- Tamil, Telugu, Gadaba, and Brahui -- as examples. Neither the phenomenon we discussed here are not exclusively Dravidian, nor are they exhaustive.  We hope more issues will be identified (and solutions will be developed) as efforts to develop (or convert) treebanks for Dravidian languages in UD framework grow in future. 

\section*{Acknowledgements}
Sowmya Vajjala contributed to this research as an employee of the National Research Council of Canada, thereby establishing a copyright belonging to the Crown in Right of Canada, that is, to the Government of
Canada. 

\bibliography{mybib}
\bibliographystyle{acl}

\end{document}